%% file: main.tex
\documentclass{article}
\ifdefined\pdfminorversion\pdfminorversion=7\fi
\usepackage{iclr2027_conference,times}
\usepackage[T1]{fontenc}
\usepackage[utf8]{inputenc}
\usepackage{amsmath,amssymb,booktabs,multirow,graphicx}
\usepackage{xurl}
\usepackage{placeins,needspace,float,wrapfig}
\usepackage{capt-of}
\usepackage{xcolor}
\usepackage{hyperref}

\makeatletter
\newsavebox{\cs@barbase}
\newsavebox{\cs@barline}
\newlength{\cs@baroffset}
\newcommand*{\xoverline}[2][0.75]{%
  \mathpalette{\cs@xoverline{#1}}{#2}%
}
\newcommand*{\cs@xoverline}[3]{%
  \begingroup
  \sbox{\cs@barbase}{$\m@th#2#3$}%
  \setbox\cs@barline\null
  \ht\cs@barline=\ht\cs@barbase
  \dp\cs@barline=\dp\cs@barbase
  \wd\cs@barline=#1\wd\cs@barbase
  \sbox{\cs@barline}{$\m@th#2\overline{\copy\cs@barline}$}%
  \setlength{\cs@baroffset}{\wd\cs@barbase}%
  \addtolength{\cs@baroffset}{-\wd\cs@barline}%
  \ifdim\wd\cs@barline<\wd\cs@barbase
    \rlap{\hskip 0.5\cs@baroffset\usebox{\cs@barline}}%
    \usebox{\cs@barbase}%
  \else
    \hskip -0.5\cs@baroffset\rlap{\usebox{\cs@barbase}}%
    \hskip 0.5\cs@baroffset\usebox{\cs@barline}%
  \fi
  \endgroup
}
\makeatother
\title{The Confidence Shortcut:\\A Reasoning Failure Mode in\\Masked Diffusion Models}
\author{
Dueun Kim$^{1}$ \quad
Albert No$^{1\,\dagger}$ \\
$^{1}$Department of Artificial Intelligence, Yonsei University
}
\iclrfinalcopy
\begin{document}
\maketitle
\begingroup
\renewcommand{\thefootnote}{\fnsymbol{footnote}}
\footnotetext[2]{Correspondence to: Albert No \texttt{\textless albertno@yonsei.ac.kr\textgreater}.}
\endgroup
\input{sections/body.tex}
\FloatBarrier
\subsection*{AI use statement}
We used generative AI tools to improve manuscript clarity and readability, refactor and debug experimental code. The authors reviewed the AI-assisted contributions incorporated into this work and take responsibility for the final manuscript, code, and reported results.

\urlstyle{same}
\bibliography{references.bib}
\bibliographystyle{iclr2027_conference}
\clearpage
\suppressfloats[t]
\input{sections/appendix.tex}
\end{document}

%% file: sections/body.tex
% NEW CITATION: wei2022chain (CoT motivation in Introduction).

\begin{abstract}

Chain-of-thought reasoning helps autoregressive models solve complex problems by generating intermediate steps that support later predictions. Masked diffusion models (MDMs) offer a similar opportunity through arbitrary-order generation: they can ideally reveal intermediate results along logical dependencies. In practice, however, standard decoding simply prioritizes high-confidence tokens, which need not align with this dependency order. We identify this discrepancy as the \emph{confidence shortcut}: models commit with high certainty to plausible tokens while neglecting long-range dependencies. In multi-digit addition, models predict higher-order digits without properly tracking carries through long chains. Controlled pretraining across diverse reasoning tasks confirms that confidence-guided ordering often selects suboptimal sequences, and confidence-aligned training schemes can exacerbate these failures---for example, increasing addition error rates by an order of magnitude. Our findings caution against relying solely on confidence to choose generation orders and against training objectives that reinforce this preference. The experimental code is available at \href{https://github.com/jinha2536/mdm-arithmetic}{this repository.}

\end{abstract}

\section{Introduction}
\label{sec:intro}

Masked diffusion models (MDMs) reconstruct masked tokens from visible context and can generate outputs in arbitrary orders \citep{sahoo2024mdlm,shi2024md4,ou2025radd}. Recent work has scaled MDMs to large language models \citep{nie2025llada,ye2025dream,bie2025llada2} and applied them to code generation, reasoning, and planning \citep{gong2026diffucoder,ye2025beyond,zhao2025d1}. Each revealed token changes the context for subsequent predictions, so a model's accuracy can depend on the order in which tokens are generated \citep{kim2025train}.

In autoregressive language models, chain-of-thought reasoning can improve accuracy by generating intermediate results to support subsequent predictions \citep{wei2022chain}. MDMs offer a related opportunity: they can reveal intermediate results before predicting tokens that depend on them. Conventional confidence-based decoding has an appealing easy-first rationale: reveal the most confident predictions, then use them to support harder ones \citep{chang2022maskgit,kim2025train}. Yet a prediction may be confident because it matches a familiar pattern, even when the dependencies determining its value remain unresolved. Does confidence actually select an order that follows logical dependencies? 

To test this, we need a task whose dependencies are known independently of the model's chosen order. In multi-digit addition, carry dependencies can be traced directly from the operands. Carries propagate from least to most significant digit \citep{lee2024teaching}, and a correct digit, together with its operand digits, makes the next carry recoverable. This provides a natural reference order that an MDM can follow while preserving most-significant-first notation. On ordinary additions, carries usually need to be traced back only a few columns, allowing a \emph{shortcut rule} based on a short operand window to predict many digits correctly. When a longer chain places the carry source beyond that window, a model relying on the rule may remain confident despite failing to resolve the dependency. We call this failure the \emph{confidence shortcut}.

In controlled pretraining experiments with standard random masking, the same trained model achieves higher accuracy under the carry order than under confidence decoding, showing that confidence-based decoding can leave avoidable errors. This limitation raises a natural concern for training: if models favor shortcut predictions, aligning training objectives with this preference risks reinforcing the failure. Indeed, recent methods such as PAPL and PUMA pursue this alignment to reduce training--inference mismatch \citep{peng2026papl,kim2026puma}. On long carry chains, both increase error rates under confidence decoding by more than an order of magnitude relative to standard random masking. Moreover, changing the decoding order alone need not restore accuracy after such confidence-aligned training.

We extend the analysis to Maze, ListOps, Countdown, and Sudoku, finding further cases of suboptimal confidence decoding and degraded reasoning under confidence-aligned training. Our contributions are threefold:
\begin{itemize}
    \item We identify the \emph{confidence shortcut}, in which models confidently commit to plausible tokens while overlooking dependencies needed for reliable reasoning.
    \item We diagnose this failure through controlled addition, showing that confidence decoding underperforms the carry order and confidence-aligned training amplifies long-carry errors.
    \item We extend the analysis across reasoning tasks, demonstrating limitations of confidence-based decoding and the risks of aligning training with a suboptimal inference policy.
\end{itemize}

\section{Preliminaries}

\label{sec:prelim}

\paragraph{Masked diffusion models.} A masked diffusion model (MDM) learns to reconstruct tokens from partially masked contexts. Given $\mathbf{x}=(x_1,\ldots,x_L)\sim p_{\text{data}}$, we sample a masking rate $\lambda\sim\mathrm{Unif}(0,1)$ and independently mask each position with probability $\lambda$. Let $M\subseteq\{1,\ldots,L\}$ be the resulting masked set, and let $\xoverline{M}=\{1,\ldots,L\}\setminus M$ denote the unmasked positions. The model observes $\mathbf{x}_{\xoverline{M}}$ and predicts a distribution over the vocabulary $\mathcal{V}$ at each masked position. The standard denoising objective is

\[
\mathcal{L}(\theta)
=
-\mathbb{E}_{\mathbf{x},\lambda,M}
\left[
\frac{1}{\lambda}
\sum_{i\in M}
\log p_\theta(x_i\mid\mathbf{x}_{\xoverline{M}})
\right],
\]

where $1/\lambda$ compensates for the smaller number of prediction targets at lower masking rates.

\paragraph{Decoding.} A policy chooses positions to reveal, and the model predicts their values. In our order comparisons, one position is revealed per step. Confidence decoding selects the masked position with the highest top-token probability \citep{chang2022maskgit},

\begin{equation*}
c_i^\theta(\mathbf{x}_{\xoverline{M}})=\max_{v\in\mathcal V}p_\theta(v\mid \mathbf{x}_{\xoverline{M}}),
\end{equation*}

then fills it with the maximizing token. Other criteria share this probability-based easy-first rationale, favoring a larger top-two probability gap (margin) or lower predictive entropy \citep{kim2025train, ye2025dream}. We focus on the widely used top-token probability rule.

\paragraph{Confidence-aligned training.}
During training, standard random masking (\textbf{Random}) masks target positions independently of the inference policy and applies equal loss weights to all masked positions. At inference, confidence decoding instead selects the most confident masked position at each step and uses its conditional distribution to generate the token. Confidence-aligned training aims to reduce this mismatch by focusing learning on the conditional predictions used during generation.

PAPL pursues this alignment by retaining random masks but increasing the cross-entropy loss weights at apparently easy positions \citep{peng2026papl}. These weights depend on the probabilities assigned to clean target tokens, rather than the maximum token probabilities used in decoding; $\alpha$ controls the strength of the additional weighting. PUMA instead aligns the partial states used for training with the confidence-selected reveal sequence \citep{kim2026puma}. It selects positions by confidence, fills them with clean tokens, and trains on the resulting states. The two methods thus pursue training--inference alignment through prediction weighting and state selection, respectively.

% \begin{figure}[t]
% \centering
% \includegraphics[width=1.0\linewidth]{figures/addition_failures.pdf}
% \caption{Completed additions from a model trained with random masking. The labels p, g, and k denote propagate, generate, and kill columns. \textbf{Top ($-1$):} a carry of 1 reaches a kill column with operand digits 5 and 1. Its correct digit is $5+1+1=7$, but the model predicts 6, corresponding to $5+1+0$ with the column's local outgoing carry of 0. \textbf{Bottom ($+1$):} a carry of 0 reaches a generate column with operand digits 9 and 5. Its correct digit is $(9+5+0)\bmod10=4$, but the model predicts 5, corresponding to $(9+5+1)\bmod10$ with its local outgoing carry of 1.}
% \label{fig:addition_errors}
% \end{figure}

\begin{figure}[t]
\centering
\includegraphics[width=1.0\linewidth]{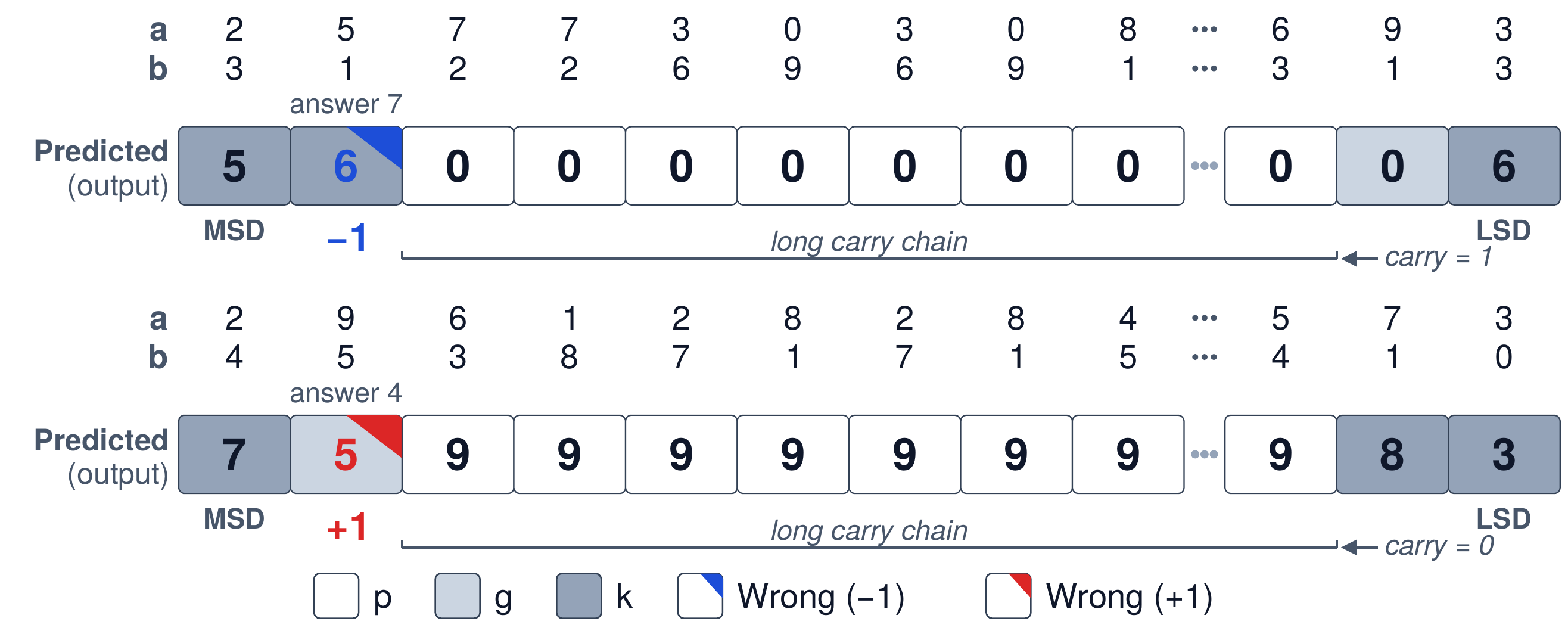}
\caption{Completed additions from a model trained with random masking. The labels p, g, and k denote propagate, generate, and kill columns. \textbf{Top ($-1$):} a carry of 1 reaches a kill column with operand digits 5 and 1. Its correct digit is $5+1+1=7$, but the model predicts 6, corresponding to $5+1+0$ with the column's local outgoing carry of 0. \textbf{Bottom ($+1$):} a carry of 0 reaches a generate column with operand digits 9 and 5. Its correct digit is $(9+5+0)\bmod10=4$, but the model predicts 5, corresponding to $(9+5+1)\bmod10$ with its local outgoing carry of 1.}
\label{fig:addition_errors}
\end{figure}

\section{Diagnosing the confidence shortcut in addition}

\label{sec:addition}

We analyze confidence decoding at the step-by-step level, asking whether each token is predicted before or after the dependencies governing its value are revealed. In multi-digit addition, these dependencies can be traced directly from the operands. Carries propagate from lower to higher digit positions, making least-significant-digit-first (LSD-first) decoding a natural reference order.

Addition also lets us separate dependency span from sequence length. A carry chain passes information from a lower column through consecutive columns to a higher answer digit. We vary chain length while keeping operand and answer lengths fixed, then evaluate the same trained model under both confidence and LSD-first decoding to compare their reveal orders and accuracies.

\subsection{Carry dependencies and LSD-first decoding}
\label{sec:carry}

We write each operand with 32 digits and predict the full 33-digit sum. Let $a_i$ and $b_i$ denote operand digits and $c_i$ the answer digit, indexed from least to most significant. With $r_i$ denoting the carry entering column $i$,

\begin{equation}
\begin{aligned}
r_0 &= 0, \qquad c_{32}=r_{32},\\
c_i &= (a_i+b_i+r_i)\bmod 10,\\
r_{i+1} &= 1\{a_i+b_i+r_i\ge10\}, \qquad i=0,\ldots,31.
\end{aligned}
\label{eq:addition}
\end{equation}

Let $s_i=a_i+b_i$ denote the local column sum excluding $r_i$. A column \textbf{kills} ($k$) the carry when $s_i\le8$, setting $r_{i+1}=0$, or \textbf{generates} ($g$) a carry when $s_i\ge10$, setting $r_{i+1}=1$, regardless of $r_i$. When $s_i=9$, it \textbf{propagates} ($p$) the incoming carry unchanged: $r_{i+1}=r_i$.

A \emph{carry chain} is a maximal run of consecutive $p$ columns; its length is the number of propagating columns. Its incoming carry is set by the nearest lower $k$ or $g$ column, or by $r_0=0$ at the least significant boundary. The non-propagating operand column immediately above the run is the \textbf{chain head}. In $g\,p\,p\,p\,k$ written most significant first, the three $p$ columns form the chain, the rightmost $k$ is its source, and the leftmost $g$ is its head. A run reaching the highest operand column instead determines the overflow digit.

The head's local role determines its outgoing carry, but its answer digit depends on the incoming carry transmitted through the chain. A \emph{shortcut rule} inspecting only $w$ lower columns can recover this carry if any column within the window determines it ($g$ or $k$). For a digit with more than $w$ lower columns, the probability that the entire window propagates is $10^{-w}$ under independent uniform operand digits. The carry source then lies beyond the window. Such a rule can therefore appear reliable on ordinary additions while failing on long carry chains.

LSD-first decoding generates digits in the order $c_0,\ldots,c_{32}$. Each correctly decoded digit $c_i$ with $i<32$, together with $a_i$ and $b_i$, makes $r_{i+1}$ recoverable, providing information for the next prediction. With correct digit predictions, this order resolves long chains through successive local calculations.

\input{tables/table_addition.tex}

\subsection{Confidence versus the carry order}

\label{sec:confidence_order}

We train a two-layer, 0.35M-parameter Transformer with standard random masking on 20,000 additions. Evaluation uses 10,000 ordinary additions and 500 examples at each minimum carry-chain length stratum. Each model is evaluated under both confidence and LSD-first decoding. Both policies use the model's predictions; LSD-first supplies no correct digits or carries.

The model trained with random masking reaches perfect accuracy on ordinary additions and achieves 1.000 exact match across all tested carry strata under LSD-first decoding. Confidence decoding nevertheless falls to 0.992 on chains of length at least 28 (Table~\ref{tab:addition_controlled}). Both policies use the same trained model: confidence-based order selection introduces errors even though following the carry dependencies lets this small model solve all tested cases.

\subsection{Effects of confidence-aligned training}

\label{sec:controlled}

\paragraph{Confidence aligned training increases long chain errors.} We next test whether confidence-aligned training amplifies long-carry errors under confidence decoding, training both PUMA and PAPL with the same data, architecture, and update budget as Random to isolate training effect.

PUMA trains on partial answers formed by selecting positions by confidence and filling clean tokens. It reaches 99.9\% exact match on ordinary additions. On chains of length at least 28, however, its confidence-decoding error rises to 9.2\%, versus 0.8\% for Random. LSD-first still solves every tested carry stratum with the same PUMA model, highlighting confidence's blind spot (Table~\ref{tab:addition_controlled}).

PAPL retains random masks but upweights masked-token losses where the model assigns higher probability to the clean target. We test $\alpha=1$ and $5$, within the effective range reported by \citet{peng2026papl}. At $\alpha=1$, it reaches 99.9\% on ordinary additions, but only 0.026 under confidence and 0.030 under LSD-first on chains of length at least 28, failing under hard dependencies. At $\alpha=5$, it also fails on ordinary additions.

\paragraph{Failures concentrate systematically at chain heads.}We examine Random and PUMA on long carry chains ending before the overflow column (which lacks operand digits). For both models, every observed failure under confidence decoding in this subset is a \emph{chain-head error} with a systematic pattern. The error signs reveal a consistent shortcut: the model substitutes the head column's locally determined outgoing carry for the incoming carry transmitted along the chain.

Specifically, at a \emph{kill} head with an incoming carry of $1$, the correct answer digit is $s_i+1$. Substituting its outgoing carry of $0$ in place of the incoming carry yields $s_i$, producing a $-1$ error (Figure~\ref{fig:addition_errors}, top). Conversely, at a \emph{generate} head with an incoming carry of $0$, the correct answer digit is $s_i-10$. Substituting its outgoing carry of $1$ in place of the incoming carry yields $s_i-10+1$, producing a $+1$ error (Figure~\ref{fig:addition_errors}, bottom). In each case, the head's local operand digits determine its outgoing carry, but they cannot resolve the incoming carry that is required to predict the answer digit correctly.

\begin{wrapfigure}{r}{0.49\textwidth}
\vspace{-0.5em}
\centering
\includegraphics[width=\linewidth]{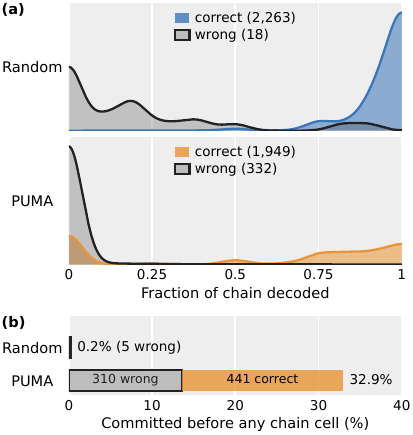}
\caption{Chain head commit timing under confidence decoding. \textbf{(a)} Density of the fraction of chain digits already decoded when the chain head is committed, for correct and incorrect predictions. \textbf{(b)} Percentage of chain-head digits committed before \emph{any} chain digit.}
\vspace{0.5em}
\label{fig:commit_timing}
\end{wrapfigure}

\paragraph{The confidence shortcut drives premature commitment at chain heads.}
The chain head's answer digit depends on the carry propagated through the chain, yet incorrect predictions are often committed while much of that chain remains masked. This already occurs under random-mask training: at an incorrect commitment, the median masked fraction is 82\% (Figure~\ref{fig:commit_timing}(a)). On chains of length at least 28, Random's median decoding step of an incorrect commitment is 10.5.

Confidence-aligned training with PUMA amplifies this pattern. Its median masked fraction at an incorrect commitment rises to 100\%, while the median step on chains of length at least 28 falls to 3. PUMA also commits the chain head before committing any carry-chain digit in 32.9\% of evaluated cases, compared with 0.2\% for Random (Figure~\ref{fig:commit_timing}(b)). All 751 such PUMA predictions follow the shortcut rule: assume an incoming carry of 1 at a generate head and 0 at a kill head. All 441 predictions whose local guess matches the actual incoming carry are correct; all 310 with a mismatched guess are wrong. These early predictions thus follow the local carry guess and invariably fail when it differs from the actual incoming carry.
\WFclear

\paragraph{Remasking during generation does not reliably correct errors.}
Generation-time confidence need not identify erroneous committed tokens \citep{zhang2025cdlm}. To test whether dynamic revision can resolve mistakes, we evaluate ReMDM schedules that vary remasking probabilities, target low-confidence commitments, or repeat remasking at fixed noise levels \citep{wang2025remasking}, alongside PRISM's learned token-quality signal \citep{kim2026prism}. On the longest carry chains, errors persist under PUMA across these methods, while PAPL shows negligible improvement (Appendix~\ref{app:remasking}).

Exact leave-one-out (LOO) correction evaluates all positions in a completed answer by masking each digit individually and scoring its current value given the remaining output. This requires 33 masked model queries per round just to select a revision site. Among the tested methods, only this costly procedure nearly restores PUMA's accuracy; PAPL remains inaccurate. However, repairing a completed answer is distinct from our central question: whether decoding can follow the task's logical order during generation, without remasking or post-hoc correction.

\subsection{The same pattern in a pretrained diffusion LLM}

\label{sec:llada}

We next test whether the same decoding failure appears in LLaDA-8B Base, an LLM-scale MDM pretrained on a 2.3-trillion-token corpus \citep{nie2025llada}. On 200 additions with carry-chain length 28, confidence decoding fails on 8.5\% of examples, whereas LSD-first solves all of them with the same predictor. Its errors likewise include highly confident, off-by-one chain head predictions while intermediate digits remain masked (Appendix~\ref{app:llada}). The confidence shortcut observed in the small controlled models thus also appears in an LLM-scale model, even on this basic arithmetic task.
\section{Reasoning across dependency structures}

\label{sec:domains}

We now ask whether confidence decoding follows the dependencies of other reasoning tasks, and whether confidence-aligned training worsens failures to follow this reasoning order. Within each task, Random, PUMA, and PAPL use the same complete training examples, architecture, and training iteration budget. PAPL uses $\alpha=5$ throughout this section. See Appendix~\ref{app:training} for data specifications and training details for each domain.

\input{tables/table_maze_main.tex}

\subsection{Maze: spatial connectivity}

\label{sec:maze}

Maze completion assigns each open cell a binary label indicating whether it lies on the unique start-to-goal path in a tree maze \citep{ivanitskiy2023maze}. This simple output can depend on connections to distant cells. A \textbf{corridor} is a maximal consecutive segment of this path excluding endpoints and branching cells; its length counts cells, including bends. Inside a long corridor, a short local view may reach neither endpoint, making a true path segment resemble an off-path passage. A shortcut rule that judges path membership from this local view can therefore confuse the two.

As in addition, a known solution procedure provides a reference reveal order. A dead end other than the start or goal cannot be part of the solution path and can be safely removed. Repeatedly removing dead ends, including those exposed by earlier removals, leaves exactly the solution path. \textbf{Dead-end filling (DEF)} reveals cells in this elimination sequence, followed by the remaining path. The input maze determines the schedule, while the model predicts each label. 

\begin{figure}[t]
\centering
\includegraphics[width=0.9\linewidth]{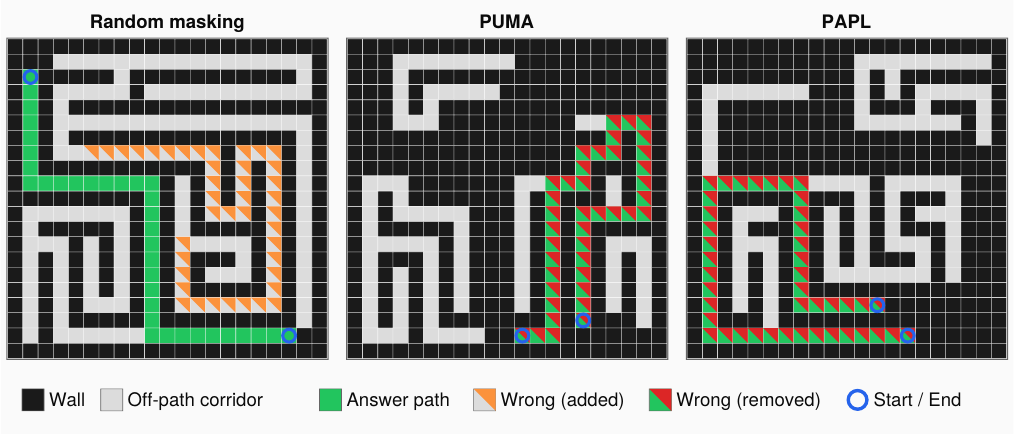}
\caption{Wrong-commit region geometry on the maze grid.
Cells are labeled by the color: walls (black), off-path corridors (light gray), the answer path (green) connecting two start/end cells (blue circles). Wrong commits are overlaid as a diagonal half: an off-path cell labeled on-path (orange) or an on-path cell labeled off-path (red).}
\label{fig:maze_errors}
\end{figure}

We train a three-layer, 1.3M-parameter Transformer on 10,000 mazes. We evaluate puzzle-level exact match on 300 mazes per corridor-length stratum, averaging results over three training seeds. On the long-corridor tail of the same generator, at minimum corridor length 60, confidence decoding achieves 0.930 for Random, 0.560 for PAPL, and 0.675 for PUMA. DEF raises these values to 1.000, 1.000, and 0.985, respectively (Table~\ref{tab:maze_main}). Confidence-aligned training thus increases errors that the same models largely avoid under the reference order.

Among the analyzed confidence-decoding failures, all incorrect cells form a single continuous path in 80\% of cases. Errors also cluster locally: 96.8\% of the open (non-wall) cells directly adjacent to an incorrect cell are themselves incorrect. The error direction differs across models. PUMA and PAPL remove segments of the true path in 97\% and 100\% of their analyzed failures, respectively, whereas Random adds an off-path branch in 73\% of its failures (Figure~\ref{fig:maze_errors}).

Post-completion LOO correction does not meaningfully improve accuracy on the long-corridor stratum (Appendix~\ref{app:additional}). Each query masks one cell while keeping the other predicted labels visible. When errors span adjacent cells, masking one leaves neighboring wrong labels in place, which may continue to support the same mistaken prediction.

\subsection{ListOps: nested computation}

\label{sec:listops}

ListOps evaluates nested expressions whose operations each return a digit from $0$ to $9$ \citep{nangia2018listops}. \texttt{SUMMOD} returns the sum of its arguments modulo $10$. The outermost operation is the \textbf{root}; \textbf{depth} counts operations along the longest nesting chain. Operations whose arguments are all input digits occupy \textbf{layer} 1; each remaining operation lies one layer above its highest-layer child operation. For \texttt{SUMMOD(MAX(MIN(2,5),7),MIN(3,8))}, both \texttt{MIN}s occupy layer 1 and return 2 and 3, \texttt{MAX} occupies layer 2 and returns 7, and the root occupies layer 3 and returns 0. The expression has depth 3 despite its shorter right branch.

Following work that includes intermediate results \citep{ballout2024show}, we predict every operation's value, storing child results before parents and the left subtree before the right: \texttt{2730} in this example. Exact match requires the complete sequence to be correct. Confidence decoding can fill these positions in any order. \textbf{Layered decoding} completes each layer before the next, using confidence within a layer: positions 1 and 3, then 2, then 4. Each parent is revealed with its child predictions visible; if these are correct, only its own operation remains.

We train an eight-layer, 11.9M-parameter Transformer on 200,000 expressions of depths 1--5 and evaluate 5,000 in-range expressions alongside separate tests at unseen depths. At depth 5, layered decoding achieves 0.996 for Random, versus 0.478 for PAPL and 0.782 for PUMA; at depth 6, one level beyond training, the corresponding accuracies fall to 0.330, 0.193, and 0.183 (Table~\ref{tab:listops}). Unlike LSD-first and DEF, layered decoding does not consistently improve on confidence decoding. Nevertheless, confidence-aligned training still degrades generalization to deeper expressions.

\input{tables/table_listops.tex}

\subsection{Countdown: constructing a solution}
\label{sec:countdown}

\input{tables/table_countdown_sudoku.tex}

ListOps supplies an expression to evaluate; Countdown requires constructing one. Given four numbers and a target, the model generates an arithmetic chain, with each operation consuming two available numbers and returning its result to the pool \citep{gandhi2024stream, ye2025beyond}. For example, 86, 28, 13, and 31 reach 96 through \texttt{86+28=114}, \texttt{31-13=18}, and \texttt{114-18=96}.

A puzzle can admit several valid chains, so we report \textbf{valid solution rate}: a chain must consume available operands without reuse, perform correct arithmetic, and end at the target. Using every input number is not required. We stratify by solution multiplicity $m$, the number of recorded valid chains; higher multiplicity generally makes puzzles easier to solve \citep{katz2025seemingly}.

We train a twelve-layer, 17.8M-parameter Transformer on 400,000 puzzles. Across 1,000 held-out test puzzles, confidence decoding achieves an overall valid solution rate of 0.530 for Random, 0.518 for PAPL, and 0.317 for PUMA. PUMA's deficit is especially pronounced on high-multiplicity inputs ($m\ge11$), where it scores 0.364 compared with 0.981 for both Random and PAPL (Table~\ref{tab:countdown_sudoku}(a)). PAPL retains random masks and reweights losses, whereas PUMA concentrates training on confidence-selected partial answers. This difference in training contexts may contribute to PUMA's difficulty constructing valid chains.

A \textbf{step-sequential} reference completes one equation before proceeding to the next, while leaving the model to choose operands and operations. For PUMA, its overall valid solution rate is 0.314, close to 0.317 under confidence decoding. Although step-sequential decoding raises PUMA's score to 0.537 on high-multiplicity inputs, it remains well below Random at 0.973 and PAPL at 0.967. Neither order resolves PUMA's difficulty constructing a valid chain.

To test completion with some arithmetic choices already established in advance, we randomly reveal half of a clean solution's tokens and let the same model complete the rest using confidence decoding. PUMA reaches 0.999 on high-multiplicity inputs and 0.992 overall (Table~\ref{tab:countdown_sudoku}(b)). This dramatic performance recovery suggests that PUMA can use a partially specified chain but struggles to establish consistent arithmetic choices from the puzzle alone.

\subsection{Sudoku: overlapping constraints}

\label{sec:sudoku}

We train an eight-layer, 5.3M-parameter Transformer on approximately 570,000 puzzles from Sudoku-Extreme, a challenging benchmark of $9\times9$ sudoku puzzles \citep{wang2025hierarchical}. Each row, column, and $3\times3$ box must contain the digits 1--9 without repetition. Filling a cell restricts the remaining choices, so which cells can be resolved next depends on the current grid.

We report accuracy on blank cells, including two especially difficult subsets. The first is the top 1\% by dataset rating, which counts guesses required by the tdoku solver. The second contains puzzles with at least 95\% of blank cells at technique level 4 (TL4): These cells remain unresolved by the reference solver's implemented logical techniques. The solver then uses search, tentatively committing to a candidate value, propagating its consequences, and backtracking when a contradiction arises. Both are solver-based difficulty measures, computed independently of model predictions.

PUMA improves accuracy under confidence decoding, including on both difficult subsets (Table~\ref{tab:countdown_sudoku}(c)). Its overall advantage over Random is 12.4 and 3.4 percentage points
under confidence and random-order decoding, respectively. Confidence decoding also exceeds the tested solver-derived cell orders in overall accuracy (Appendix~\ref{app:reference_orders}). Here, PUMA improves an already effective decoding policy, illustrating that the effects of confidence alignment depend on the task.

\section{Discussion}
\label{sec:discussion}

Our results highlight the confidence shortcut: models commit confidently to plausible tokens without adequately accounting for the dependencies that determine their values. Local shortcut rules that succeed on common cases can sustain this confidence even on harder cases where they fail. Confidence-based decoding may favor these predictions over an order that reveals useful intermediate results, and confidence-aligned training can reinforce the same preference.

\paragraph{Better orders can reveal existing capabilities.}
Reference orders derived from task structure substantially improve accuracy for the same trained model. For example, LSD-first decoding enables PUMA to solve long-carry additions, while dead-end filling (DEF) boosts Maze performance. Because the reference order only dictates when to predict each token, the model itself must still compute the correct intermediate values. These gains thus demonstrate that the model already possesses reasoning capabilities that confidence decoding fails to elicit. Consequently, improving order selection can unlock these capabilities without modifying model parameters.

\paragraph{Confidence alignment can limit recovery.}
Training around an imperfect order-selection heuristic introduces a further risk. While alignment benefits tasks like Sudoku where confident predictions provide reliable constraints, it may harm broader reasoning. On long-carry addition, PAPL remains inaccurate even under LSD-first decoding; on ListOps, both PAPL and PUMA underperform Random. The benefits of confidence-aligned training are therefore highly task-dependent, and its distortions cannot always be undone by switching to a reference order at inference.

\paragraph{Random masking preserves useful alternatives.}
Random masking may seem inefficient by training on contexts a specific decoder rarely visits. However, this broader coverage provides opportunities to learn
predictions needed by alternative decoding orders, such as predicting higher digits when lower ones are already revealed. Random's advantage across carry chains, long corridors, and ListOps demonstrates that broad, uniformly weighted supervision supports reliable predictions on structurally difficult cases. Our findings thus support retaining random-mask pretraining as a robust foundation while focusing on improving decoding order.

\paragraph{Difficult examples and reference orders can guide learning.}
A promising direction is to train or guide models using reference orders from known solution procedures, encouraging them to compute and leverage intermediate results when local shortcuts fail. Such order selection can be learned post-hoc \citep{hong2026upo,xu2026sas} or end-to-end \citep{ma2025recor,huang2025dcolt}. Crucially, candidate orders must be evaluated on hard dependency structures rather than assumed to be sound, ensuring that models preserve the underlying computations required for robust reasoning.

\section{Conclusion}

\label{sec:conclusion}

Confidence is often useful for generation, but it can favor premature shortcut rules that overlook reasoning dependencies. Confidence-aligned training can reinforce these shortcuts, and the tested reference orders do not always restore accuracy. Our findings support retaining random-mask training as a foundation while developing and evaluating more reliable decoding orders for each task.

%% file: tables/table_addition.tex
\begin{table}[t]
\centering
\setlength{\belowcaptionskip}{3pt}
\caption{Addition exact match under confidence and LSD-first decoding. PUMA's long-carry accuracy recovers under LSD-first, while substantial errors remain after PAPL training.}
\label{tab:addition_controlled}
\begingroup
\small
\renewcommand{\arraystretch}{1.12}
\setlength{\tabcolsep}{4pt}
\begin{tabular}{rcccccccc}
\toprule
\multirow{2}{*}{Min. chain length} & \multicolumn{2}{c}{Random} & \multicolumn{2}{c}{PAPL ($\alpha=1$)} & \multicolumn{2}{c}{PUMA} & \multicolumn{2}{c}{PAPL ($\alpha=5$)} \\
\cmidrule(lr){2-3}\cmidrule(lr){4-5}\cmidrule(lr){6-7}\cmidrule(lr){8-9}
& Conf. & LSD & Conf. & LSD & Conf. & LSD & Conf. & LSD \\
\midrule
Standard & 1.000 & 1.000 & 0.999 & 0.999 & 0.999 & 1.000 & 0.000 & 0.000 \\
\midrule
4  & 0.998 & 1.000 & 0.914 & 0.942 & 0.868 & 1.000 & 0.000 & 0.000 \\
12 & 0.996 & 1.000 & 0.618 & 0.724 & 0.864 & 1.000 & 0.000 & 0.000 \\
20 & 0.984 & 1.000 & 0.462 & 0.580 & 0.826 & 1.000 & 0.000 & 0.000 \\
24 & 0.994 & 1.000 & 0.254 & 0.308 & 0.868 & 1.000 & 0.000 & 0.000 \\
28 & 0.992 & 1.000 & 0.026 & 0.030 & 0.908 & 1.000 & 0.000 & 0.006 \\
\bottomrule
\end{tabular}
\endgroup
\end{table}

%% file: tables/table_maze_main.tex
\begin{table}[t]
\centering
\setlength{\belowcaptionskip}{3pt}
\caption{Maze exact match under confidence and dead-end filling (DEF), stratified by the longest corridor on the solution path. DEF largely recovers accuracy on the long-corridor stress strata.}
\label{tab:maze_main}
\begingroup
\small
\renewcommand{\arraystretch}{1.12}
\setlength{\tabcolsep}{4pt}
\begin{tabular}{rcccccc}
\toprule
\multirow{2}{*}{Min. corridor} & \multicolumn{2}{c}{Random} & \multicolumn{2}{c}{PAPL} & \multicolumn{2}{c}{PUMA} \\
\cmidrule(lr){2-3}\cmidrule(lr){4-5}\cmidrule(lr){6-7}
& Conf. & DEF & Conf. & DEF & Conf. & DEF \\
\midrule
4  & 0.993 & 0.990 & 0.903 & 0.995 & 0.871 & 0.916 \\
8  & 0.994 & 0.990 & 0.903 & 0.990 & 0.861 & 0.902 \\
15 & 0.977 & 0.974 & 0.893 & 0.978 & 0.920 & 0.896 \\
20 & 0.990 & 0.943 & 0.888 & 0.968 & 0.803 & 0.888 \\
25 & 0.986 & 0.944 & 0.892 & 0.988 & 0.821 & 0.923 \\
30 & 0.988 & 0.950 & 0.898 & 0.993 & 0.870 & 0.934 \\
\midrule
\multicolumn{7}{l}{\emph{Long corridor tail}} \\
50 & 0.930 & 1.000 & 0.690 & 0.990 & 0.645 & 0.985 \\
60 & 0.930 & 1.000 & 0.560 & 1.000 & 0.675 & 0.985 \\
\bottomrule
\end{tabular}
\endgroup
\end{table}

%% file: tables/table_listops.tex
\begin{table}[t]
\centering
\setlength{\belowcaptionskip}{3pt}
\caption{ListOps full-trace exact match under confidence and layered bottom-up decoding. A substantial training gap remains when child predictions precede their parents.}
\label{tab:listops}
\begingroup
\small
\renewcommand{\arraystretch}{1.12}
\setlength{\tabcolsep}{4pt}
\begin{tabular}{lcccccc}
\toprule
\multirow{2}{*}{Depth stratum} & \multicolumn{2}{c}{Random} & \multicolumn{2}{c}{PAPL} & \multicolumn{2}{c}{PUMA} \\
\cmidrule(lr){2-3}\cmidrule(lr){4-5}\cmidrule(lr){6-7}
& Conf. & Layer. & Conf. & Layer. & Conf. & Layer. \\
\midrule
$d= 1$ & 1.000 & 1.000 & 1.000 & 1.000 & 1.000 & 1.000 \\
$d= 2$ & 1.000 & 1.000 & 1.000 & 1.000 & 1.000 & 1.000 \\
$d= 3$ & 1.000 & 1.000 & 0.974 & 0.958 & 0.998 & 0.992 \\
$d= 4$ & 0.996 & 0.998 & 0.658 & 0.640 & 0.922 & 0.898 \\
$d= 5$ & 0.996 & 0.996 & 0.466 & 0.478 & 0.830 & 0.782 \\
\midrule
OOD: $d=6$ & 0.330 & 0.330 & 0.230 & 0.193 & 0.203 & 0.183 \\
OOD: $d=7$ & 0.013 & 0.017 & 0.033 & 0.033 & 0.010 & 0.000 \\
\bottomrule
\end{tabular}
\endgroup
\end{table}

%% file: tables/table_countdown_sudoku.tex
% Compact replacement. Requires booktabs and multirow, already loaded by main.tex.
% All 63 reported values are retained. The main manuscript is not modified.
\begin{table}[t]
\centering
\setlength{\belowcaptionskip}{3pt}
\caption{Countdown valid solution rate and Sudoku blank-cell accuracy. Panels (a) and (c) report generation from the input alone. Panel (b) fixes 50\% of a clean solution's tokens and completes the remainder by confidence decoding. Random, PAPL, and PUMA denote training schemes.}
\label{tab:countdown_sudoku}
\begingroup
\small
\renewcommand{\arraystretch}{1.05}
\setlength{\tabcolsep}{2pt}

\textbf{(a) Countdown: input-only generation}\par\smallskip
\begin{tabular*}{\linewidth}{@{\extracolsep{\fill}}l*{9}{c}@{}}
\toprule
\multirow{2}{*}{Multiplicity} & \multicolumn{3}{c}{Confidence} & \multicolumn{3}{c}{Random order} & \multicolumn{3}{c}{Step-sequential} \\
\cmidrule(lr){2-4}\cmidrule(lr){5-7}\cmidrule(lr){8-10}
& Random & PAPL & PUMA & Random & PAPL & PUMA & Random & PAPL & PUMA \\
\midrule
$m\in[1,3]$ & 0.190 & 0.173 & 0.219 & 0.138 & 0.117 & 0.098 & 0.129 & 0.081 & 0.109 \\
$m\in[4,10]$ & 0.519 & 0.494 & 0.506 & 0.487 & 0.410 & 0.355 & 0.513 & 0.481 & 0.423 \\
$m\geq11$ & 0.981 & 0.981 & 0.364 & 0.989 & 0.981 & 0.501 & 0.973 & 0.967 & 0.537 \\
Overall & 0.530 & 0.518 & 0.317 & 0.503 & 0.478 & 0.285 & 0.497 & 0.467 & 0.314 \\
\bottomrule
\end{tabular*}

\par\medskip
\begin{minipage}[t]{0.36\linewidth}
\centering
\textbf{(b) Countdown: 50\% clean}\par\smallskip
\begin{tabular*}{\linewidth}{@{\extracolsep{\fill}}lccc@{}}
\toprule
\multirow{2}{*}{Multiplicity} & \multicolumn{3}{c}{Training scheme} \\
\cmidrule(lr){2-4}
& Random & PAPL & PUMA \\
\midrule
$m\in[1,3]$ & 0.995 & 0.993 & 0.985 \\
$m\geq11$ & 1.000 & 1.000 & 0.999 \\
Overall & 0.997 & 0.998 & 0.992 \\
\bottomrule
\end{tabular*}
\end{minipage}\hfill
\begin{minipage}[t]{0.62\linewidth}
\centering
\textbf{(c) Sudoku}\par\smallskip
\begin{tabular*}{\linewidth}{@{\extracolsep{\fill}}l*{6}{c}@{}}
\toprule
\multirow{2}{*}{Stratum} & \multicolumn{3}{c}{Confidence} & \multicolumn{3}{c}{Random order} \\
\cmidrule(lr){2-4}\cmidrule(lr){5-7}
& Random & PAPL & PUMA & Random & PAPL & PUMA \\
\midrule
Overall & 0.629 & 0.630 & 0.753 & 0.565 & 0.570 & 0.599 \\
Top-1\% rating & 0.357 & 0.372 & 0.609 & 0.370 & 0.365 & 0.432 \\
TL4 frac. $\geq0.95$ & 0.407 & 0.404 & 0.539 & 0.382 & 0.376 & 0.414 \\
\bottomrule
\end{tabular*}
\end{minipage}
\endgroup
\end{table}

%% file: sections/appendix.tex
% Appendix draft refined on 2026-09-26 from the supplied ICLR source.
% Hardware, runtimes, and ListOps architecture updated from author confirmation.

\appendix

\section{Related work}
\label{app:related}

\paragraph{Masked diffusion models.}
Discrete diffusion models generate discrete data by learning to reverse a token-corruption process~\citep{austin2021structured,campbell2022continuous,lou2024sedd}. Masked diffusion specializes this process to masking and learns to reconstruct missing tokens from visible context~\citep{sahoo2024mdlm,shi2024md4}. \citet{ou2025radd} connect absorbing diffusion to any-order autoregressive modeling. Recent models extend this framework to large-scale language modeling and code generation~\citep{nie2025llada,ye2025dream,gong2026diffucoder}, as well as reasoning and planning~\citep{ye2025beyond}.

\paragraph{Generation order.}
\citet{chang2022maskgit} prioritize high-probability predictions. \citet{kim2025train} study margin-based adaptive ordering that avoids difficult conditionals, and \citet{ye2025dream} introduce entropy-based ordering. \citet{cai2026confidence} analyze the efficiency of entropy-based decoding under exact conditional marginals. For structural guidance, \citet{aman2026logicdiff} uses logic-role classifiers, \citet{asano2026where} train planners from ordering oracles, and \citet{st2026think} conditions generation on autoregressive plans. \citet{peng2025p2} separate planning from token prediction in P2. \citet{hong2026upo} and \citet{xu2026sas} learn unmasking policies with frozen denoisers. In contrast, \citet{hong2026lomdm} jointly learn denoising and generation order, as do \citet{ma2025recor} and \citet{huang2025dcolt}. \citet{zhao2025d1} and \citet{wang2026d2} improve diffusion reasoning through reinforcement learning of the model itself. Learning an order with a frozen denoiser changes which conditionals are queried, while joint training can change their accuracy.

\paragraph{Confidence-aligned training.}
\citet{peng2026papl} introduce PAPL, which adds planner-dependent loss weights to random-mask training. \citet{kim2026puma} instead train PUMA on confidence-selected, teacher-forced partial states. Both aim to align training with the inference policy. Their changes to training contexts or loss weights also relate to curriculum learning within each example~\citep{Bengiocurriculum}. We examine whether confidence alignment reinforces shortcut rules and weakens predictions needed by alternative reasoning orders.

\paragraph{Reasoning dependencies and shortcuts.}
\citet{geirhos2020shortcut} describe shortcut learning as reliance on correlated cues that can fail under distribution shift. Algorithmic tasks expose such reliance through explicit dependencies. In arithmetic, formatting, intermediate supervision, and positional embeddings affect generalization~\citep{lee2024teaching,mcleish2024transformers}; Sudoku and maze studies examine search and recurrent computation~\citep{shah2024causal,wang2025hierarchical}. \citet{nangia2018listops} introduce ListOps to test recursive composition, and \citet{ballout2024show} add intermediate subexpression results. \citet{jeong2026carry} study carry representations, while \citet{hong2026logical} examine consistency across equivalent presentations. We use tasks with explicit computations to test whether confidence follows the dependencies that determine an answer.

\paragraph{Remasking and correction.}
\citet{zhang2025cdlm} find that standard MDM confidence is unreliable for locating errors in code revision and improve correction by supervising visible corruptions. \citet{kim2026prism} learn token-quality scores through self-correction in PRISM. Our correction experiments test whether additional model-based signals can repair highly confident reasoning errors.

\FloatBarrier

\section{Data construction and experimental protocol}
\label{app:training}

\subsection{Shared setup}

Within each task, Random, PAPL, and PUMA use the same complete training examples, architecture, and update budget. The prompt remains visible throughout training and decoding; only answer positions are masked and predicted. Evaluation strata are defined by the generator or solver annotations, independently of model predictions. We report accuracy within each structural stratum and separately evaluate unstratified Addition inputs.

The controlled models are pre-norm transformers with bidirectional self-attention, GELU activations, learned absolute positional embeddings, tied input and output projections, and an MLP width three times the embedding width. Table~\ref{tab:training_settings} lists the architecture, learning rate, update count, and PUMA schedule for each task. All models use batch size 256 and AdamW with $\beta=(0.9,0.95)$, weight decay $0.01$, and gradient clipping at $1.0$. Training runs for the prescribed number of updates without early stopping. Each run uses a single A100 GPU and takes approximately four hours for Addition and Maze and ten hours for ListOps, Countdown, and Sudoku. The capacity ablation increases model size while holding the data and update count fixed.

Unless stated otherwise, decoding reveals one position per step and fills it with the model's highest-probability token. Confidence selects the position with the highest confidence score, random order samples a masked position uniformly, and the reference policies use the task procedures described below. The model parameters are unchanged across these order comparisons. Sampling and correction experiments use the alternative procedures in Appendix~\ref{app:additional}.

\paragraph{Reference-order comparisons.}
LSD-first, dead-end filling, and layered bottom-up decoding use known calculation or elimination procedures to choose positions, while the same trained model predicts every revealed value. These comparisons test whether following the corresponding dependencies improves accuracy.

\input{tables/table_training_settings.tex}

\subsection{PAPL: reweighting masked-token losses}

PAPL~\citep{peng2026papl} retains random-mask sampling and changes the relative weight of masked-token losses. For a nonempty mask set $M$, the within-state weights are

\begin{equation*}
\begin{aligned}
\widetilde w_i^\theta
&=\frac{\exp\!\left(\log p_\theta(x_i\mid \mathbf{x}_{\xoverline{M}})/\tau\right)}
{\sum_{j\in M}\exp\!\left(\log p_\theta(x_j\mid \mathbf{x}_{\xoverline{M}})/\tau\right)},\\
w_i^\theta
&=\frac{1+\alpha\widetilde w_i^\theta}{|M|},\qquad i\in M.
\end{aligned}
\end{equation*}

The per-position cross-entropy is multiplied by $1+\alpha\widetilde w_i^\theta$ before averaging over masked positions. Every position retains a base weight, with an additional contribution determined by its clean-token probability. Setting $\alpha=0$ recovers uniform weighting within the mask set.

The planner uses the probability of the clean target token, which is already available during loss computation. This differs from decoding confidence $c_i^\theta(\mathbf{x}_{\xoverline{M}})$: the latter takes a maximum over vocabulary tokens and does not require the correct answer. We use $\tau=1$ throughout, with $\alpha=1$ for Addition and $\alpha=5$ for Maze, ListOps, Countdown, and Sudoku. In the Addition ablation, $\alpha=5$ yields zero exact match under confidence decoding at every tested minimum chain length (4, 12, 20, 24, and 28). Under LSD-first, accuracy is also zero except at minimum length 28, where it is $0.006$. The main Addition comparison therefore uses $\alpha=1$.

\subsection{PUMA: training on confidence trajectories}

PUMA~\citep{kim2026puma} changes which partial states are used for training. Starting with a masked answer, it ranks the remaining positions by confidence and reveals a group of high-confidence positions. The values placed at those positions are always the clean tokens. Thus, the trajectory follows the model's position choices while remaining consistent with the training answer.

\paragraph{Streaming buffer.}

We maintain one teacher-forced chain per batch slot. Each update trains on the current partial states and then advances each chain by one reveal stage. A completed chain is replaced by a new example with its answer fully masked. The same forward pass provides both the token losses and the confidence scores used to advance the chains. This reuses intermediate states across updates instead of rebuilding a complete trajectory for every training example at every update.

\paragraph{Reveal granularity.}

The number of stages $K$ controls how many tokens are revealed together. For an answer region of length $L$, a stage reveals roughly $L/K$ tokens, with mild randomization of the reveal count. Larger $K$ gives finer trajectories that more closely approach one-token-at-a-time decoding. The schedule increases $K$ from its start value to its end value during the first third of training and keeps it fixed thereafter. The endpoints appear in Table~\ref{tab:training_settings}. For Addition, $K$ increases from 3 to 16 over a 33-position answer, giving roughly two revealed digits per stage at the end of the schedule.

\subsection{Addition}

\paragraph{Format and generation.}

The operands are sampled independently and uniformly from $\{0,\ldots,10^{32}-1\}$, and the answer is their full sum $a+b$. Each operand is zero-padded to 32 digits; the sum uses 33 digits to accommodate overflow. With the least-significant-first indices of Equation~\ref{eq:addition}, the displayed sequence is

\begin{equation*}
\underbrace{a_{31}\cdots a_0\;\texttt{+}\;b_{31}\cdots b_0\;\texttt{=}}_{\text{prompt: 66 positions}}
\underbrace{c_{32}\cdots c_0}_{\text{answer: 33 positions}}.
\end{equation*}

Thus, mathematical indices increase with significance, while the text uses the usual most-significant-first order. The four-digit analogue of $47+38$ is \texttt{0047+0038=00085}. The vocabulary contains the ten digits, \texttt{+}, \texttt{=}, a mask token, and a padding token. All 33 answer digits, including leading zeros and the possible overflow digit, enter exact-match evaluation.

\paragraph{LSD-first reference order.}
Ordinary column addition starts at the units digit and passes any carry to the next column. LSD-first decoding follows this direction~\citep{lee2024teaching}, revealing the answer from $c_0$ through $c_{32}$. A correct immediately lower answer digit, together with its operands, determines the next carry:
\begin{equation*}
r_i=\mathbf{1}\{a_{i-1}+b_{i-1}>c_{i-1}\},\qquad 1\le i\le32.
\end{equation*}
For example, when the lower operand digits sum to nine, an answer digit of zero indicates carry one, whereas an answer digit of nine indicates carry zero. The next prediction can therefore recover its incoming carry from the immediately lower column, rather than tracing it through a longer sequence of propagating operand columns. Carries are not supplied explicitly, and the visible answer digits are model predictions.

\paragraph{Carry strata and evaluation sets.}

A column propagates its incoming carry when $a_i+b_i=9$. Carry-chain length is measured by the longest consecutive run of such columns. Training uses 20,000 uniformly sampled additions. The standard evaluation contains 10,000 examples from the same operand distribution; the controlled carry sweep contains 500 examples per minimum-length stratum. Figure~\ref{fig:commit_timing} pools the minimum-length strata 4, 12, 20, 24, and 28, using 2,281 cases per model for the commit-timing analysis. Correct and incorrect commit-timing densities are each normalized to unit area. Appendix~\ref{app:controlled_timing} gives the commitment analysis.

\subsection{Maze}

\paragraph{Generation and evaluation.}

The generator follows the maze-completion setting of \citet{ivanitskiy2023maze}. Starting from a random cell, randomized depth-first search generates a connected tree maze on a $10\times10$ logical grid. In the rendered $21\times21$ grid, walls separate logical cells except where the search opens a passage. Start and goal form a maximally separated pair. Training uses 10,000 mazes. The backbone is the unique start-to-goal path; branching points on it have at least three accessible neighbors. A corridor is a maximal consecutive segment of the backbone excluding start, goal, and branching cells; its length counts cells along the path, including bends. The structural coordinate is the maximum corridor length in the maze. The evaluation pool contains 5,000 mazes, with 300 examples per stratum at minimum corridor lengths 4, 8, 15, 20, 25, and 30. We additionally evaluate minimum corridor lengths of 50 and 60. Corridors of at least 50 and 60 cells occur in approximately $14.3\%$ and $6.1\%$ of generated mazes, respectively. These strata therefore select the long-corridor tail of the same generator. All strata report accuracy means over three training seeds.

\paragraph{Dead-end-filling reference order.}
A dead end has only one passage connecting it to the remaining maze. Unless it is the start or goal, it cannot belong to the path between them: a path entering it would have to turn back. Removing that cell preserves the solution and can make its neighbor a new dead end. Repeating this step removes side branches from their tips toward the remaining maze. In a tree maze, every cell outside the unique start-to-goal path is eventually removed, leaving exactly that path.

Dead-end filling (DEF) computes this pruning sequence from the input maze before generation. The decoder reveals cells in successive elimination rounds, followed by the surviving path from start to goal. It uses the resulting schedule throughout generation, independently of the labels predicted by the model. During pruning, the next position is selected by a rule that safely eliminates off-path cells, while the model must still assign the labels correctly.

\paragraph{Sequence format.}

Each example has the form \texttt{puzzle=solution}. Both grids contain $21\times21=441$ cells and are serialized in row-major order. The prompt uses \texttt{\#} for walls, \texttt{.} for unmarked corridor cells, and \texttt{S}/\texttt{E} for the endpoints. The answer preserves walls and endpoints and labels the remaining open cells as \texttt{1} on the unique start-to-goal path and \texttt{0} off the path. The following $3\times3$ logical maze, rendered as a $7\times7$ grid, illustrates the input and output:
\begin{center}
\begin{tabular}{c@{\qquad}c}
Prompt & Answer \\
\texttt{\#\#\#\#\#\#\#} & \texttt{\#\#\#\#\#\#\#} \\
\texttt{\#S....\#} & \texttt{\#S1100\#} \\
\texttt{\#\#\#.\#\#\#} & \texttt{\#\#\#1\#\#\#} \\
\texttt{\#.....\#} & \texttt{\#00111\#} \\
\texttt{\#\#\#\#\#.\#} & \texttt{\#\#\#\#\#1\#} \\
\texttt{\#E....\#} & \texttt{\#E1111\#} \\
\texttt{\#\#\#\#\#\#\#} & \texttt{\#\#\#\#\#\#\#}
\end{tabular}
\end{center}
Line breaks are shown only for readability; each grid is stored as a flat string. Puzzle-level exact match requires the complete predicted answer grid to be correct.

\subsection{ListOps}

\paragraph{Task and output.}

The original ListOps task predicts the final value of a nested expression~\citep{nangia2018listops}. Explained-ListOps adds intermediate subexpression results to the generated target~\citep{ballout2024show}. We likewise predict one digit for every operation. The outermost operation is the \emph{root}, and expression \emph{depth} is the longest chain of nested operations. Values are stored in post-order, visiting children from left to right and recording each operation's value after those of its children, without operator names. Evaluation requires the complete value sequence to be correct.

\paragraph{Generation and evaluation.}
We recursively generate expression trees using MIN, MAX, MEDIAN, and SUM-MOD-10. At each internal node, the operator is sampled uniformly and the child count follows a fixed distribution concentrated at two to four children; leaf digits are sampled uniformly from 0 to 9. Training uses 200,000 expressions with target depths from 1 to 5, sampled so that deeper expressions are less frequent. All objectives use the same corpus and padding format. In-range evaluation uses 1,000 expressions per depth stratum. Depths 6 and 7 are held out from training.

\paragraph{Layered bottom-up reference order.}
The first layer contains operations whose arguments are input digits. Each remaining operation is assigned one layer above the highest layer of its internal children. These layers are fixed by the expression tree before decoding. Every position in a layer is revealed before any position in the next layer, with the model's score selecting among positions within the current layer.

For example, \(\mathrm{SUMMOD}(\mathrm{MAX}(\mathrm{MIN}(2,5),7),\mathrm{MIN}(3,8))\) has depth 3, with a shorter right branch. Its post-order values are \texttt{2730}: the left MIN gives 2, its MAX parent gives 7, the right MIN gives 3, and the root gives 0. Layer 1 contains output positions 1 and 3; layer 2 contains position 2; layer 3 contains position 4. Following these layers makes every child's predicted value visible before its parent is selected. If those values are correct, the parent requires only its own operation. The model predicts all intermediate values; post-order specifies where the values are stored, whereas the layered schedule specifies when they are revealed.

\paragraph{Sequence format.}

Operators are encoded as \texttt{X} (MAX), \texttt{N} (MIN), \texttt{D} (MEDIAN), and \texttt{S} (SUM-MOD-10). The prompt is a prefix-bracketed expression, and the answer contains one digit for each internal node in post-order. For example,

\begin{equation*}
\underbrace{\texttt{[S [X [N 2 5] 7] [N 3 8]]=}}_{\text{prompt}}
\underbrace{\texttt{2730}}_{\text{answer}}.
\end{equation*}

This is the same expression used in the layered example above. Spaces separate expression elements for readability. The stored order does not force confidence decoding to reveal a child before its parent: the position for 7 can be selected before the position for 2.

\subsection{Countdown}

\paragraph{Source and evaluation.}

We use the four-number Countdown data and equation-chain format of \citet{ye2025beyond}. Each clean answer consists of three binary operations, using addition, subtraction, multiplication, or division, and ends at the requested target. Solution multiplicity $m$ counts distinct valid chains recorded by the dataset annotator; a larger $m$ means that more chains reach the target. The shared training corpus contains approximately 400,000 puzzles, with low-multiplicity examples subsampled so that puzzles with more recorded solutions are more common during training. Evaluation uses 1,000 puzzles from the source distribution, stratified into $m\in[1,3]$, $m\in[4,10]$, and $m\ge11$.

We report \emph{strict valid solution rate} (VSR). Starting with the multiset of four input numbers, the evaluator checks each generated equation, consumes its two operands from the available pool, and inserts its correctly computed result. A valid chain passes these operand-consumption and arithmetic checks and has the target as its final result. The evaluator does not require every input number to be used or the final pool to contain only the target. Malformed equations, unavailable operands, incorrect calculations, and a wrong final target receive no credit. Duplicate input values are tracked with their multiplicities. Any chain satisfying these conditions is accepted, irrespective of the annotated answer. Annotated-chain exact match gave similar aggregate results; we report strict VSR to credit alternative valid chains.

\paragraph{Sequence format.}

The prompt lists four input numbers and a target. The answer is a comma-separated chain of three binary operations:

\begin{equation*}
\begin{aligned}
\text{prompt:}\quad&\texttt{86,28,13,31,96=}\\
\text{answer:}\quad&\texttt{86+28=114,31-13=18,114-18=96}.
\end{aligned}
\end{equation*}

Each operation consumes two available numbers and adds its result to the remaining pool. The final result must equal the target.

\paragraph{Step-sequential decoding.}
\label{app:countdown-extra}
This schedule completes one equation at a time, using boundaries from the clean solution string. Within an equation, operand and operator characters are revealed from left to right, followed by the equality sign, result, and comma before the next equation. Padding follows the answer positions. The schedule supplies the answer format; the model still predicts every token and chooses the arithmetic operations needed to reach the target.

\paragraph{Completion from part of a clean solution.}
For a clean solution with $L$ non-padding tokens, we sample $\lfloor L/2\rfloor$ positions uniformly without replacement and keep their clean values fixed. The unchanged model fills the remaining positions by confidence decoding. The completed chain is evaluated with the same validity check; a valid completion need not reproduce the unrevealed part of the clean solution. The revealed tokens provide parts of the arithmetic choices and intermediate results, allowing the model to complete a partially specified solution.

\subsection{Sudoku}

\paragraph{Source and evaluation.}

We use a fixed mixture drawn from the 3,831,994-puzzle \texttt{sudoku-extreme} corpus~\citep{wang2025hierarchical}, with fewer puzzles retained from higher solver-rating tiers. Ratings count guesses made by the tdoku solver~\citep{tdoku}. The resulting training set contains approximately 570,000 puzzles and is shared across objectives. Reported accuracy is restricted to originally blank cells. Accuracy is evaluated in rating bins 0, 1--17, 18--34, 35--51, 52--101, and $\ge102$, with 1,000, 1,000, 500, 200, 100, and 500 test puzzles, respectively. Its top-1\% tier therefore refers to 500 puzzles rated at least 102. Rating strata and cell-level solving-technique strata are separate annotations. The TL4 fraction is the proportion of originally blank cells still unresolved by the reference solver's implemented logical techniques.

\paragraph{Sequence format and technique labels.}

The puzzle and solution are flattened in row-major order. The prompt contains the given digits and \texttt{.} for blank cells; the answer is the complete 81-digit solution. The sequence is \texttt{puzzle=solution}.

Technique levels group solving rules documented by SudokuWiki~\citep{sudokuwiki} and describe the procedure used by the reference solver to establish a cell's value. TL0 covers naked singles, where a cell has only one candidate, and hidden singles, where a digit has only one possible location in a row, column, or box. TL1 uses subsets and intersections to eliminate candidates from other cells. TL2 uses fish and wing patterns to link candidate restrictions across cells, and TL3 uses forcing chains to follow the consequences of tentative assignments. TL4 marks cells left unresolved by these implemented techniques.

\paragraph{Constraint propagation and position selection.}
If a row contains the digits 1 through 8, its remaining cell must be 9. Placing that digit also removes 9 as a candidate elsewhere in the same column and box, potentially making other cells easier to determine. As placements change the remaining candidates, they also change which cells can be resolved next. Confidence decoding uses the model's current predictions to choose the next cell, following the intuition of filling apparently determined cells first.

\paragraph{Solver-derived orders.}
\label{app:sudoku-extra}
\label{app:reference_orders}
The \emph{constraint-propagation order} follows a solver that repeatedly fills naked and hidden singles. When no single remains, it selects the cell with the fewest candidates, breaks ties in row-major order, and tries candidate digits in ascending order. Each guess is followed by further propagation; a contradiction triggers backtracking. Cells filled in one propagation stage are recorded in row-major order, and assignments from failed branches are removed from the sequence. The \emph{technique-difficulty order} instead sorts cells by their assigned level, from TL0 to TL4, using discovery order within each level. Both schedules are computed before generation, and the model predicts the blank-cell values in the resulting order.

In DEF, selecting a dead end also establishes that it is off the solution path. A Sudoku solver's next cell may instead be selected for a trial assignment whose success is determined only by later propagation and search. Transferring that cell sequence gives the model a schedule, but it must still infer the values that make the solution consistent. Overall blank-cell accuracies are $0.578/0.581/0.631$ for constraint-propagation order and $0.581/0.584/0.624$ for technique-difficulty order, for Random/PAPL/PUMA, respectively. Confidence decoding reaches $0.629/0.630/0.753$ (Table~\ref{tab:countdown_sudoku}), supporting its usefulness for selecting the next cell in this task.

\subsection{Variable-length answers}
\label{app:rainbow}

ListOps and Countdown have variable-length answers. For batched training, the answer region is padded to a fixed length using a deterministic cyclic sequence of distinct padding tokens, following Rainbow Padding~\citep{kim2026rainbow}. The answer regions have 20 positions for ListOps and 40 for Countdown. All training objectives within a task use the same answer format and padding scheme. The task examples above display the answer content before padding.

\FloatBarrier

\section{Addition evaluation and error analysis}
\label{app:addition_errors}

\subsection{LLaDA-8B evaluation}
\label{app:llada}

We evaluate LLaDA-8B Base~\citep{nie2025llada} without updating its parameters. The prompt contains four demonstrations, followed by the test addition. The demonstrations include sums with and without overflow, so the leading digit of the 33-digit answer can be either 1 or 0. Confidence, LSD-first, and random-order decoding use the same examples and model, revealing one answer position per step.

The evaluation contains 200 ordinary additions sampled without conditioning on carry-chain length and 200 examples at each planted chain length 4, 8, 20, 24, and 28. Table~\ref{tab:llada_addition} reports exact match over the complete answer. At chain length 28, confidence decoding solves 183 of 200 examples, whereas LSD-first solves all 200. Fifteen of the 17 failed confidence-decoded answers contain an incorrect chain-head digit. These incorrect chain-head predictions have mean probability $0.96$ and are committed with $51\%$ of the chain digits still masked on average.

\input{tables/table_llada.tex}

\subsection{Controlled-model commitment analysis}
\label{app:controlled_timing}

Figure~\ref{fig:commit_timing} pools the minimum carry-chain lengths 4, 12, 20, 24, and 28 from Table~\ref{tab:addition_controlled}, excluding chains that reach the overflow column. The analysis contains 2,281 cases per model: 2,263 correct and 18 incorrect cases for Random, and 1,949 correct and 332 incorrect cases for PUMA. All 350 failures include an incorrect chain-head prediction.

For each case, we record the fraction of chain digits already decoded when the chain head is committed. Zero means that no chain digit is visible; one means the full chain is visible. Correct and incorrect cases are separately normalized to unit-area densities in Figure~\ref{fig:commit_timing}(a). Among incorrect commitments, the median fraction still masked is $82\%$ for Random and $100\%$ for PUMA. On the minimum-length-28 stratum, the median incorrect-commitment step is 10.5 for Random and 3 for PUMA.

An early commitment predicts the chain head before any digit in its carry chain is revealed. Such commitments account for $0.2\%$ of Random cases and $32.9\%$ of PUMA cases (Figure~\ref{fig:commit_timing}(b)). At a non-propagating chain head with operand sum $s_i=a_i+b_i$, the shortcut rule guesses incoming carry 1 at a generate column and 0 at a kill column:
\begin{equation*}
\widetilde r_i=\mathbf{1}\{s_i\ge10\},\qquad
\widetilde c_i=(s_i+\widetilde r_i)\bmod10.
\end{equation*}
All 751 early PUMA predictions follow this rule. All 441 predictions whose local guess matches the actual incoming carry are correct; all 310 with a mismatched guess are wrong. These early predictions therefore follow the local carry guess and invariably fail when it differs from the actual incoming carry.

\FloatBarrier
\section{A carry-frontier probe}
\label{app:probe}

We test whether a frozen model's hidden states identify digits whose incoming carry is locally available. A masked position is eligible when this carry can be determined from the immediately lower operand column, using its correct visible answer digit when needed. The units position is eligible because $r_0=0$; the tens position is eligible from the units operands. At higher positions, the lower column must kill or generate a carry, or its correct answer digit must already be visible. These eligible positions form the \emph{carry frontier}.

For each model, we fit a linear score on standardized final hidden vectors to rank eligible positions above ineligible positions within the same partial answer. The model parameters remain fixed. Fitting uses random masks, LSD-first prefixes, and confidence-selected prefixes, always filled with clean visible digits. Controlled-model fitting uses 100 ordinary additions and 60 each at chain lengths 8, 20, and 28, separate from the evaluation examples. LLaDA fitting uses 40 examples each at lengths 4, 8, 12, and 16, with the four-shot setup in Appendix~\ref{app:llada}.

During generation, the probe selects the highest-scoring masked position and the unchanged model predicts the digit. Each predicted digit then becomes visible context. We evaluate 60 additions at each chain length 8, 20, and 28 for every model. Probe scheduling raises PUMA from 141/180 to 180/180 and LLaDA from 165/180 to 179/180, while reducing accuracy for Random and PAPL (Table~\ref{tab:probe_rollouts}). A learned score for local carry availability can therefore improve position selection, but its benefit depends on the model.

\input{tables/table_probe_rollouts.tex}
\FloatBarrier

\section{Capacity, sampling, and correction}
\label{app:additional}

\subsection{Capacity}

Increasing capacity tests whether the long-carry gap reflects insufficient model size (Table~\ref{tab:addition_capacity}). Data size and update count remain fixed. The base model has 2 layers, 2 heads, and width 128 (0.35M parameters); the larger model has 6 layers, 6 heads, and width 384 (8.92M, approximately $26\times$ larger). Base-model rows average three training seeds.

\input{tables/table_capacity.tex}

\subsection{Sampling cost}

Sampling can recover a correct answer that greedy prediction misses, but may require many complete generations. We therefore ask how many PUMA samples are needed to match one completion from a Random-trained model. We evaluate a PUMA checkpoint from training seed 42 at 300{,}000 updates on 300 additions with minimum carry-chain length 28. For each problem, we draw 128 samples at each temperature $T\in\{1.0,1.5,2.0\}$, without top-$p$ truncation. We estimate pass@$k$ from per-problem success counts~\citep{chen2021evaluating}. A problem counts as solved whenever an oracle finds a correct candidate among the $k$ samples.

\begin{figure}[t]
\centering
\includegraphics[width=0.8\linewidth]{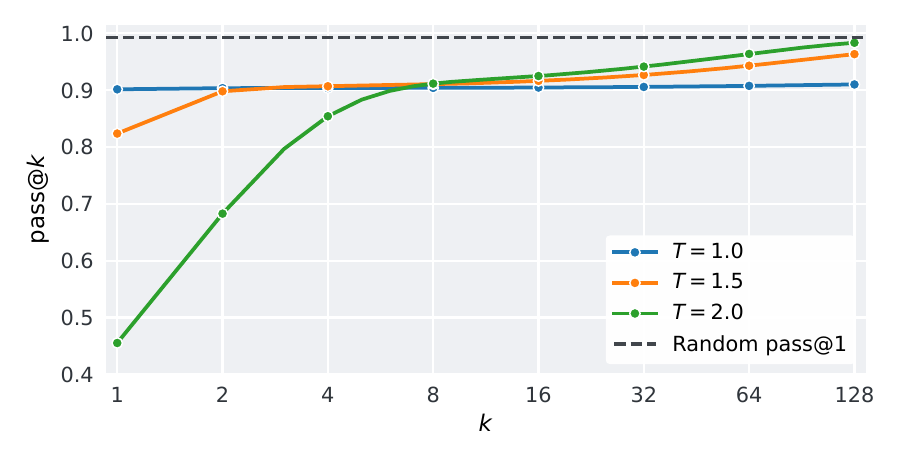}
\caption{PUMA pass@$k$ on Addition with minimum carry-chain length 28. The dashed Random pass@1 line is the confidence-decoding reference ($0.992$; Table~\ref{tab:addition_capacity}).}
\label{fig:addition_sampling_cost}
\end{figure}

The Random-trained model reaches $99.2\%$ with one confidence-decoded completion in the three-seed reference evaluation (Table~\ref{tab:addition_capacity}). PUMA's highest pass@$k$ across the tested temperatures rises from $90.15\%$ at $k=1$ to $96.37\%$ at $k=64$ and $98.33\%$ at $k=128$ (Figure~\ref{fig:addition_sampling_cost}). None of the observed curves reaches the reference accuracy within 128 candidates. Sampling recovers many errors, but requires a substantially larger generation budget to approach the accuracy available from random-mask training in one completion. Here, cost counts generated candidates; it excludes temperature search and verification.

\subsection{Remasking}
\label{app:remasking}

Remasking reopens committed tokens for prediction. ReMDM-cap uses a common, capped remasking probability; ReMDM-conf favors tokens with lower confidence at their last reveal; and ReMDM-loop repeatedly remasks and predicts at a fixed noise level~\citep{wang2025remasking}. PRISM instead selects revisions using a learned token-quality head that scores the current partial answer~\citep{kim2026prism}.

\paragraph{Post-completion correction.}
Re-noising masks part of a completed answer and generates those values again. Exact leave-one-out (LOO) correction masks each digit of a completed confidence-decoded answer in turn and scores the probability of its current value given the remaining digits. Each round uses 33 queries, revises the lowest-scoring digit by argmax, and recomputes scores after a change. It stops when the selected digit is unchanged or after 33 rounds.

\input{tables/table_remasking.tex}

At minimum carry-chain length 28, PUMA improves from $0.901$ exact match to $0.909$--$0.931$ with ReMDM and $0.951$ with PRISM, leaving errors under every tested generation-time method. Post-completion re-noising reaches $0.965$, while exact LOO reaches $0.999$ (Table~\ref{tab:remasking}). Among these methods, near-complete recovery requires post-completion LOO and its additional queries. PAPL remains at $0.027$--$0.031$ on this stratum.

\paragraph{Maze.}
For Maze, LOO applies the same selection and stopping rule to open cells of a completed answer, keeping walls and endpoints fixed; the round cap is the largest eligible-cell count in the batch. At minimum corridor length 60, exact match remains $0.930$ for Random and $0.560$ for PAPL, and changes from $0.675$ to $0.685$ for PUMA. Substantial errors remain, whereas DEF reaches $1.000$, $1.000$, and $0.985$, respectively.

\FloatBarrier

%% file: tables/table_training_settings.tex
\begin{table}[htbp]
\centering
\caption{Architectures and optimization settings shared across objectives within each task. $K$ denotes the number of PUMA reveal stages.}
\label{tab:training_settings}
\begingroup
\small
\renewcommand{\arraystretch}{1.12}
\setlength{\tabcolsep}{5pt}
\begin{tabular}{lrrrr}
\toprule
Task & Layers & Heads & Width & Parameters \\
\midrule
Addition & 2 & 2 & 128 & 0.35M \\
Maze & 3 & 3 & 192 & 1.29M \\
ListOps & 8 & 8 & 384 & 11.9M \\
Countdown & 12 & 12 & 384 & 17.79M \\
Sudoku & 8 & 8 & 256 & 5.32M \\
\bottomrule
\end{tabular}
\par\medskip
\begin{tabular}{lccc}
\toprule
Task & Updates & Learning rate & PUMA $K$: start $\to$ end \\
\midrule
Addition & 300,000 & $10^{-3}$ & $3\to16$ \\
Maze & 50,000 & $3\times10^{-4}$ & $10\to40$ \\
ListOps & 300,000 & $3\times10^{-4}$ & $2\to10$ \\
Countdown & 200,000 & $3\times10^{-4}$ & $4\to20$ \\
Sudoku & 300,000 & $3\times10^{-4}$ & $8\to40$ \\
\bottomrule
\end{tabular}
\endgroup
\end{table}

%% file: tables/table_llada.tex
\begin{table}[t]
\centering
\setlength{\belowcaptionskip}{3pt}
\caption{LLaDA-8B addition exact match by reveal order and carry stratum. Standard denotes additions sampled without conditioning on carry-chain length.}
\label{tab:llada_addition}
\begingroup
\small
\renewcommand{\arraystretch}{1.12}
\setlength{\tabcolsep}{4pt}
\begin{tabular*}{\linewidth}{@{\extracolsep{\fill}}lcccccc@{}}
\toprule
Reveal order & Standard & Chain 4 & Chain 8 & Chain 20 & Chain 24 & Chain 28 \\
\midrule
Confidence & 0.995 & 0.955 & 0.870 & 0.975 & 0.945 & 0.915 \\
LSD-first  & 1.000 & 0.995 & 0.995 & 1.000 & 0.995 & 1.000 \\
Random     & 0.950 & 0.870 & 0.560 & 0.530 & 0.485 & 0.485 \\
\bottomrule
\end{tabular*}
\endgroup
\end{table}

%% file: tables/table_probe_rollouts.tex
\begin{table}[htbp]
\centering
\caption{Full-answer exact match under confidence and probe-based position selection. The same model predicts every digit. Each evaluation has 60 examples at each chain length 8, 20, and 28.}
\label{tab:probe_rollouts}
\begin{tabular}{lcc}
\toprule
Model / training & Confidence & Probe \\
\midrule
Controlled / Random & 175/180 & 167/180 \\
Controlled / PAPL & 74/180 & 43/180 \\
Controlled / PUMA & 141/180 & 180/180 \\
LLaDA-8B Base & 165/180 & 179/180 \\
\bottomrule
\end{tabular}
\end{table}

%% file: tables/table_capacity.tex
\begin{table}[H]
\centering
\caption{Addition exact match after increasing model capacity. PUMA's confidence-decoding gap persists on long carry chains.}
\label{tab:addition_capacity}
\begingroup
\small
\renewcommand{\arraystretch}{1.12}
\setlength{\tabcolsep}{4pt}
\begin{tabular}{llccccc}
\toprule
\multirow{2}{*}{Training / order} & \multirow{2}{*}{Capacity} & \multicolumn{5}{c}{Minimum carry-chain length} \\
\cmidrule(lr){3-7}
& & 4 & 12 & 20 & 24 & 28 \\
\midrule
\multirow{2}{*}{Random / confidence} & Base & 0.998 & 0.996 & 0.984 & 0.994 & 0.992 \\
& $26\times$ & 1.000 & 1.000 & 1.000 & 1.000 & 0.996 \\
\midrule
\multirow{2}{*}{PUMA / confidence} & Base & 0.868 & 0.864 & 0.826 & 0.868 & 0.908 \\
& $26\times$ & 0.988 & 0.910 & 0.844 & 0.746 & 0.730 \\
\midrule
\multirow{2}{*}{PUMA / LSD-first} & Base & 1.000 & 1.000 & 1.000 & 1.000 & 1.000 \\
& $26\times$ & 0.990 & 0.954 & 0.956 & 0.962 & 0.970 \\
\bottomrule
\end{tabular}
\endgroup
\end{table}

%% file: tables/table_remasking.tex
\begin{table}[htbp]
\centering
\caption{Addition exact match at minimum carry-chain length 28. Base uses deterministic confidence decoding; $\dagger$ marks correction after completion.}
\label{tab:remasking}
\begin{tabular}{lccc}
\toprule
Decoding / correction & Random & PAPL & PUMA \\
\midrule
Base & 0.991 & 0.027 & 0.901 \\
ReMDM-cap & 0.991 & 0.027 & 0.911 \\
ReMDM-conf & 0.992 & 0.028 & 0.909 \\
ReMDM-loop & 0.993 & 0.028 & 0.931 \\
PRISM & 0.993 & 0.028 & 0.951 \\
Re-noise$^{\dagger}$ & 0.997 & 0.029 & 0.965 \\
Exact LOO$^{\dagger}$ & 1.000 & 0.031 & 0.999 \\
\bottomrule
\end{tabular}
\end{table}

%% file: references.bib
@article{aman2026logicdiff,
  title={LogicDiff: Logic-Guided Denoising Improves Zero-Shot Reasoning in Masked Diffusion Language Models},
  author={Aman, Shaik},
  journal={arXiv:2603.26771},
  year={2026}
}

@article{asano2026where,
  title={Where-to-unmask: Ground-truth-guided unmasking order learning for masked diffusion language models},
  author={Asano, Hikaru and Kozuno, Tadashi and Saito, Kuniaki and Baba, Yukino},
  journal={arXiv:2602.09501},
  year={2026}
}

@inproceedings{austin2021structured,
  title = {Structured Denoising Diffusion Models in Discrete State-Spaces},
  author = {Austin, Jacob and Johnson, Daniel D. and Ho, Jonathan and Tarlow, Daniel and van den Berg, Rianne},
  booktitle = {NeurIPS},
  year = {2021}
}

@inproceedings{ballout2024show,
  title={Show me how it’s done: the role of explanations in fine-tuning language models},
  author={Ballout, Mohamad and Krumnack, Ulf and Heidemann, Gunther and K{\"u}hnberger, Kai-Uwe},
  booktitle={ACML},
  year={2024},
}

@inproceedings{Bengiocurriculum,
  title = {Curriculum Learning},
  author = {Bengio, Yoshua and Louradour, J{\'e}r{\^o}me and Collobert, Ronan and Weston, Jason},
  booktitle = {ICML},
  year = {2009}
}

@article{bie2025llada2,
  title={Llada2.0: Scaling up diffusion language models to 100b},
  author={Bie, Tiwei and Cao, Maosong and Chen, Kun and Du, Lun and Gong, Mingliang and Gong, Zhuochen and Gu, Yanmei and Hu, Jiaqi and Huang, Zenan and Lan, Zhenzhong and others},
  journal={arXiv:2512.15745},
  year={2025}
}

@article{cai2026confidence,
  title={Confidence-based decoding is provably efficient for diffusion language models},
  author={Cai, Changxiao and Li, Gen},
  journal={arXiv:2603.22248},
  year={2026}
}

@inproceedings{
campbell2022continuous,
title={A Continuous Time Framework for Discrete Denoising Models},
author={Andrew Campbell and Joe Benton and Valentin De Bortoli and Tom Rainforth and George Deligiannidis and Arnaud Doucet},
booktitle={NeurIPS},
year={2022},
}

@inproceedings{chang2022maskgit,
  title={Maskgit: Masked generative image transformer},
  author={Chang, Huiwen and Zhang, Han and Jiang, Lu and Liu, Ce and Freeman, William T},
  booktitle={CVPR},
  year={2022},
}

@article{chen2021evaluating,
  title={Evaluating large language models trained on code},
  author={Chen, Mark and Tworek, Jerry and Jun, Heewoo and Yuan, Qiming and Pinto, Henrique Ponde De Oliveira and Kaplan, Jared and Edwards, Harri and Burda, Yuri and Joseph, Nicholas and Brockman, Greg and others},
  journal={arXiv:2107.03374},
  year={2021}
}

@inproceedings{
gandhi2024stream,
title={Stream of Search (SoS): Learning to Search in Language},
author={Kanishk Gandhi and Denise H J Lee and Gabriel Grand and Muxin Liu and Winson Cheng and Archit Sharma and Noah Goodman},
booktitle={COLM},
year={2024},
}

@article{geirhos2020shortcut,
  title={Shortcut learning in deep neural networks},
  author={Geirhos, Robert and Jacobsen, J{\"o}rn-Henrik and Michaelis, Claudio and Zemel, Richard and Brendel, Wieland and Bethge, Matthias and Wichmann, Felix A},
  journal={Nature Machine Intelligence},
  year={2020},
}

@inproceedings{
gong2026diffucoder,
title={DiffuCoder: Understanding and Improving Masked Diffusion Models for Code Generation},
author={Shansan Gong and Ruixiang ZHANG and Huangjie Zheng and Jiatao Gu and Navdeep Jaitly and Lingpeng Kong and Yizhe Zhang},
booktitle={ICLR},
year={2026},
}

@inproceedings{
hong2026logical,
title={Logical Equivariance in Diffusion Language Models},
author={Sunwoo Hong and Seo Hyun Kim and Chen-Hao Chao and Younwoo Choi and Se-Young Yun and Rahul G Krishnan},
booktitle={ICML 2026 Workshop on Foundations of Deep Generative Models},
year={2026},
}

@inproceedings{
hong2026lomdm,
title={Unifying Masked Diffusion Models with Various Generation Orders and Beyond},
author={Chunsan Hong and Sanghyun Lee and Jong Chul Ye},
booktitle={ICML},
year={2026},
}

@inproceedings{
hong2026upo,
title={Improving Discrete Diffusion Unmasking Policies Beyond Explicit Reference Policies},
author={Chunsan Hong and Seonho An and Min-Soo Kim and Jong Chul Ye},
booktitle={ICLR},
year={2026},
}

@inproceedings{
huang2025dcolt,
title={Reinforcing the Diffusion Chain of Lateral Thought with Diffusion Language Models},
author={Zemin Huang and Zhiyang Chen and Zijun Wang and Tiancheng Li and Guo-Jun Qi},
booktitle={NeurIPS},
year={2025},
}

@article{ivanitskiy2023maze,
  title={A configurable library for generating and manipulating maze datasets},
  author={Ivanitskiy, Michael Igorevich and Shah, Rusheb and Spies, Alex F and R{\"a}uker, Tilman and Valentine, Dan and Rager, Can and Quirke, Lucia and Mathwin, Chris and Corlouer, Guillaume and Behn, Cecilia Diniz and others},
  journal={arXiv:2309.10498},
  year={2023}
}

@inproceedings{
jeong2026carry,
title={Masked Diffusion Training Induces Resampling-like Carry Representations in Addition},
author={Hwangyu Jeong and Chanwoo Park},
booktitle={Mechanistic Interpretability Workshop at ICML 2026},
year={2026},
}

@article{katz2025seemingly,
  title={Seemingly simple planning problems are computationally challenging: The countdown game},
  author={Katz, Michael and Kokel, Harsha and Sreedharan, Sarath},
  journal={arXiv:2508.02900},
  year={2025}
}

@inproceedings{
kim2025train,
title={Train for the Worst, Plan for the Best: Understanding Token Ordering in Masked Diffusions},
author={Jaeyeon Kim and Kulin Shah and Vasilis Kontonis and Sham M. Kakade and Sitan Chen},
booktitle={ICML},
year={2025},
}

@inproceedings{
kim2026prism,
title={Fine-Tuning Masked Diffusion for Provable Self-Correction},
author={Jaeyeon Kim and Seunggeun Kim and Taekyun Lee and David Z. Pan and Hyeji Kim and Sham M. Kakade and Sitan Chen},
booktitle={ICML},
year={2026},
}

@inproceedings{
kim2026puma,
title={Stop Training for the Worst: Progressive Unmasking Accelerates Masked Diffusion Training},
author={Jaeyeon Kim and Jonathan Geuter and David Alvarez-Melis and Sham M. Kakade and Sitan Chen},
booktitle={ICML},
year={2026},
}

@inproceedings{
kim2026rainbow,
title={Rainbow Padding: Mitigating Early Termination in Instruction-Tuned Diffusion {LLM}s},
author={BumJun Kim and Dongjae Jeon and Dueun Kim and Wonje Jeung and Albert No},
booktitle={ICLR},
year={2026},
}

@inproceedings{
lee2024teaching,
title={Teaching Arithmetic to Small Transformers},
author={Nayoung Lee and Kartik Sreenivasan and Jason D. Lee and Kangwook Lee and Dimitris Papailiopoulos},
booktitle={ICLR},
year={2024},
}

@inproceedings{lou2024sedd,
  title = {Discrete Diffusion Modeling by Estimating the Ratios of the Data Distribution},
  author = {Lou, Aaron and Meng, Chenlin and Ermon, Stefano},
  booktitle = {ICML},
  year = {2024}
}

@inproceedings{
ma2025recor,
title={Reinforced Context Order Recovery for Adaptive Reasoning and Planning},
author={Long Ma and Fangwei Zhong and Yizhou Wang},
booktitle={NeurIPS},
year={2025},
}

@inproceedings{
mcleish2024transformers,
title={Transformers Can Do Arithmetic with the Right Embeddings},
author={Sean Michael McLeish and Arpit Bansal and Alex Stein and Neel Jain and John Kirchenbauer and Brian R. Bartoldson and Bhavya Kailkhura and Abhinav Bhatele and Jonas Geiping and Avi Schwarzschild and Tom Goldstein},
booktitle={The 4th Workshop on Mathematical Reasoning and AI at NeurIPS'24},
year={2024},
}

@inproceedings{nangia2018listops,
  title={Listops: A diagnostic dataset for latent tree learning},
  author={Nangia, Nikita and Bowman, Samuel},
  booktitle={NAACL Student Research Workshop},
  year={2018}
}

@inproceedings{
nie2025llada,
title={Large Language Diffusion Models},
author={Shen Nie and Fengqi Zhu and Zebin You and Xiaolu Zhang and Jingyang Ou and Jun Hu and JUN ZHOU and Yankai Lin and Ji-Rong Wen and Chongxuan Li},
booktitle={NeurIPS},
year={2025},
}

@inproceedings{
ou2025radd,
title={Your Absorbing Discrete Diffusion Secretly Models the Conditional Distributions of Clean Data},
author={Jingyang Ou and Shen Nie and Kaiwen Xue and Fengqi Zhu and Jiacheng Sun and Zhenguo Li and Chongxuan Li},
booktitle={ICLR},
year={2025},
}

@inproceedings{
peng2025p2,
title={Path Planning for Masked Diffusion Models with Applications to Biological Sequence Generation},
author={Fred Zhangzhi Peng and Zachary Bezemek and Sawan Patel and Jarrid Rector-Brooks and Sherwood Yao and Alexander Tong and Pranam Chatterjee},
booktitle={ICLR 2025 Workshop on Deep Generative Model in Machine Learning: Theory, Principle and Efficacy},
year={2025},
}

@inproceedings{peng2026papl,
  title     = {Planner Aware Path Learning in Diffusion Language Models Training},
  author    = {Peng, Fred Zhangzhi and Bezemek, Zachary and Rector-Brooks, Jarrid and Zhang, Shuibai and Bronstein, Michael M. and Zhang, Anru and Bose, Joey and Tong, Alexander},
  booktitle = {ICLR},
  year      = {2026}
}

@inproceedings{sahoo2024mdlm,
  title={Simple and effective masked diffusion language models},
  author={Sahoo, Subham S and Arriola, Marianne and Schiff, Yair and Gokaslan, Aaron and Marroquin, Edgar and Chiu, Justin T and Rush, Alexander and Kuleshov, Volodymyr},
  booktitle={NeurIPS},
  year={2024}
}

@inproceedings{
shah2024causal,
title={Causal language modeling can elicit search and reasoning capabilities on logic puzzles},
author={Kulin Shah and Nishanth Dikkala and Xin Wang and Rina Panigrahy},
booktitle={NeurIPS},
year={2024},
}

@inproceedings{
shi2024md4,
title={Simplified and Generalized Masked Diffusion for Discrete Data},
author={Jiaxin Shi and Kehang Han and Zhe Wang and Arnaud Doucet and Michalis Titsias},
booktitle={NeurIPS},
year={2024},
}

@article{st2026think,
  title={Think first, diffuse fast: Improving diffusion language model reasoning via autoregressive plan conditioning},
  author={St Sauver, Earl J},
  journal={arXiv:2603.13243},
  year={2026}
}

@misc{sudokuwiki,
  author       = {Stuart, Andrew},
  title        = {{SudokuWiki}: Strategy Families},
  year         = {2008},
  howpublished = {Online resource},
  url          = {https://www.sudokuwiki.org/Strategy_Families},
}

@misc{tdoku,
  author       = {Dillon, Tom},
  title        = {{Tdoku}: A Fast {Sudoku} Solver and Generator},
  year         = {2019},
  howpublished = {GitHub repository},
  url          = {https://github.com/t-dillon/tdoku},
}

@article{wang2025hierarchical,
  title={Hierarchical Reasoning Model},
  author={Guan Wang and Jin Li and Yuhao Sun and Xing Chen and Chang-Le Liu and Yue Wu and Meng Lu and Sen Song and Yasin Abbasi Yadkori},
  journal={arXiv:2506.21734},
  year={2025},
}

@inproceedings{wang2025remasking,
title={Remasking Discrete Diffusion Models with Inference-Time Scaling},
author={Wang, Guanghan and Schiff, Yair and Sahoo, Subham and Kuleshov, Volodymyr},
booktitle={NeurIPS},
year={2025},
}

@inproceedings{
wang2026d2,
title={d2: Improving Reasoning in Diffusion Language Models via Trajectory Likelihood Estimation},
author={Guanghan Wang and Gilad Turok and Yair Schiff and Marianne Arriola and Volodymyr Kuleshov},
booktitle={ICML},
year={2026},
}

@inproceedings{wei2022chain,
  title={Chain-of-thought prompting elicits reasoning in large language models},
  author={Wei, Jason and Wang, Xuezhi and Schuurmans, Dale and Bosma, Maarten and Xia, Fei and Chi, Ed and Le, Quoc V and Zhou, Denny and others},
  booktitle={NeurIPS},  year={2022}
}

@inproceedings{
xu2026sas,
title={Scheduling Thoughts: Learning the Order of Thought in Diffusion Language Models},
author={Jiawei Xu and Minghui Liu and Aakriti Agrawal and Yifan Chen and Furong Huang},
booktitle={ICML},
year={2026},
}

@inproceedings{
ye2025beyond,
title={Beyond Autoregression: Discrete Diffusion for Complex Reasoning and Planning},
author={Jiacheng Ye and Jiahui Gao and Shansan Gong and Lin Zheng and Xin Jiang and Zhenguo Li and Lingpeng Kong},
booktitle={ICLR},
year={2025},
}

@article{ye2025dream,
  title={Dream 7B: Diffusion Large Language Models},
  author={Ye, Jiacheng and Xie, Zhihui and Zheng, Lin and Gao, Jiahui and Wu, Zirui and Jiang, Xin and Li, Zhenguo and Kong, Lingpeng},
  journal={arXiv:2508.15487},
  year={2025}
}

@article{zhang2025cdlm,
  title={Corrective diffusion language models},
  author={Zhang, Shuibai and Peng, Fred Zhangzhi and Zhang, Yiheng and Pan, Jin and Chrysos, Grigorios G},
  journal={arXiv:2512.15596},
  year={2025}
}

@inproceedings{
zhao2025d1,
title={d1: Scaling Reasoning in Diffusion Large Language Models via Reinforcement Learning},
author={Siyan Zhao and Devaansh Gupta and Qinqing Zheng and Aditya Grover},
booktitle={NeurIPS},
year={2025},
}
